\documentclass[letterpaper, 10 pt, conference]{ieeeconf}  
\IEEEoverridecommandlockouts                              
\usepackage[english]{babel}
\usepackage[T1]{fontenc}

\usepackage[nolist]{acronym}
\usepackage[cmex10]{amsmath}
\usepackage[font={small}]{caption}
\usepackage{subcaption}
\usepackage{amssymb}
\usepackage{bm}
\usepackage{booktabs}
\usepackage{color}
\usepackage{float}
\usepackage{graphicx}
\usepackage{hyperref}
\usepackage{import}
\usepackage{mathtools}
\usepackage{multirow}
\usepackage{multicol}
\usepackage{flushend}
\usepackage{outline}
\usepackage{paralist}
\usepackage{siunitx}
\usepackage{stmaryrd}
\usepackage{systeme}
\usepackage{url}
\usepackage{tikz}
\usepackage{booktabs}
\usepackage{balance}
\usepackage{cite}
\usetikzlibrary{tikzmark}
\usepackage{esvect}
\usepackage[normalem]{ulem}
\usepackage{bbm}
\usepackage{siunitx}
\usepackage{wrapfig,lipsum,booktabs}

\usepackage{caption}
\usepackage{subcaption}
\usepackage{graphicx}
\usepackage{subcaption}
\usepackage{adjustbox}
\usepackage{xcolor}
\usepackage{flushend}
\usepackage{booktabs}

\newcommand{\website}{https://zhengmaohe.github.io/leg-manip}

\usepackage{colortbl}
\definecolor{ourcolor}{HTML}{99e0eb}
\definecolor{ourblue}{HTML}{27a2c3}
\definecolor{tablecolor}{HTML}{ccf2f5} 
\definecolor{tablecolor2}{HTML}{ffcdb4}
\definecolor{citecolor}{HTML}{fe7b5b}
\definecolor{grey}{rgb}{0.9, 0.9, 0.9}

\definecolor{gred}{rgb}{0.859,0.267,0.216}
\definecolor{ggreen}{rgb}{0.059,0.616,0.345}
\definecolor{deepblue}{HTML}{27a2c3}
\definecolor{deepred}{HTML}{fe7b5b}

\newcommand{\dd}[2]{$#1\scriptstyle{\pm#2}$}
\newcommand{\ddbf}[2]{\cellcolor{tablecolor}$\mathbf{#1\scriptstyle{\pm#2}}$}

\title{\LARGE \bf Learning Visual Quadrupedal Loco-Manipulation from Demonstrations}
\author{Zhengmao He$^{1,2}$, Kun Lei$^{1}$, Yanjie Ze$^{1}$, Koushil Sreenath$^{3}$, Zhongyu Li$^{3}$, Huazhe Xu$^{1,4}$ \\
\texttt{\url{\website}}
}

\begin{document}

\twocolumn[{
    \begin{@twocolumnfalse}
    
    \maketitle
    \begin{center}
        \includegraphics[width=\textwidth, trim={0 72mm 7cm 0}, clip]{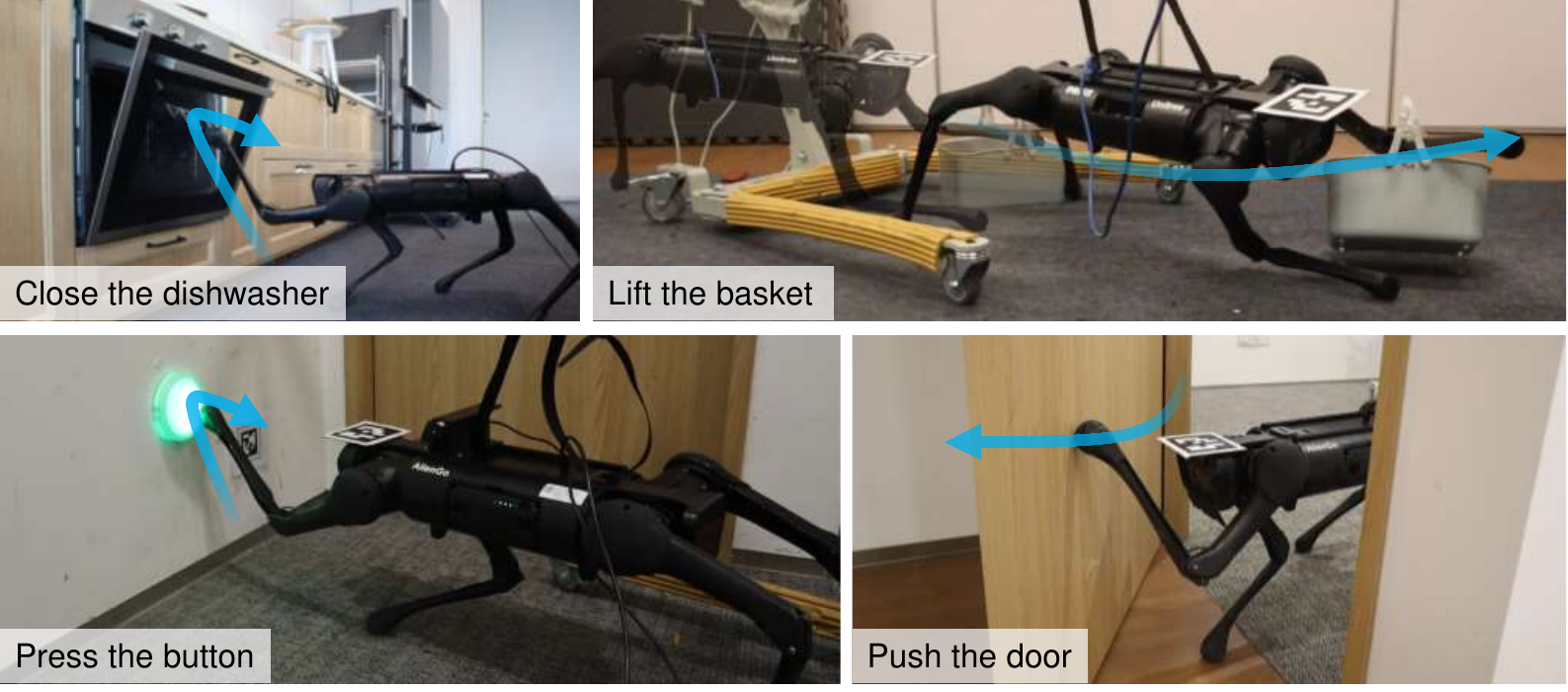}
        \captionof{figure}{We present a hierarchical learning framework to learn general loco-manipulation skills for quadruped robots. The framework enables a Unitree Aliengo robot to perform diverse skills in the real-world, including lifting baskets, pressing buttons, opening doors, and closing dishwashers, all while maintaining stable locomotion over a long distance. 
         Videos are on the~\href{\website}{project website}.
        }
        \label{fig:teaser}
    \end{center}
    \end{@twocolumnfalse}
}]

\footnotetext[1]{Shanghai Qi Zhi Institute.}
\footnotetext[2]{The Hong Kong University of Science and Technology (Guangzhou). zhe037@connect.hkust-gz.edu.cn}
\footnotetext[3]{University of California, Berkeley. \{zhongyu\_li, koushils\}@berkeley.edu}
\footnotetext[4]{Tsinghua University, IIIS. huazhe\_xu@mail.tsinghua.edu.cn}

\begin{abstract}
Quadruped robots are progressively being integrated into human environments.
Despite the growing locomotion capabilities of quadrupedal robots, their interaction with objects in realistic scenes is still limited.
While additional robotic arms on quadrupedal robots enable manipulating objects, they are sometimes redundant given that a quadruped robot is essentially a mobile unit equipped with four limbs, each possessing 3 degrees of freedom (DoFs).
Hence, we aim to empower a quadruped robot to execute real-world manipulation tasks using only its legs. 
We decompose the loco-manipulation process into a low-level reinforcement learning (RL)-based controller and a high-level Behavior Cloning (BC)-based planner. 
By parameterizing the manipulation trajectory, we synchronize the efforts of the upper and lower layers, thereby leveraging the advantages of both RL and BC.
Our approach is validated through simulations and real-world experiments, demonstrating the robot's ability to perform tasks that demand mobility and high precision, such as lifting a basket from the ground while moving, closing a dishwasher, pressing a button, and pushing a door.

\end{abstract}

\section{Introduction}
\label{sec:intro}

The four 3-DoF limbs of a quadrupedal robot, and its 6-DoF floating base torso can provide a wide workspace and flexibility for manipulation tasks. This insight underscores the inherent versatility of quadruped robots as loco-manipulators, capable of integrating locomotion and manipulation without additional robotic arms.

Previous research on using quadruped robot legs for manipulation has enabled robots to execute tasks such as pressing buttons, dribbling balls, and ball shooting~\cite{cheng2023legmanip, huang2023creating, ji2023Hierarchical,Ji2023DribbleBot}. However, these studies still face certain limitations: (\romannumeral1) designs are often task-specific, leading to poor adaptability for different tasks~\cite{cheng2023legmanip, huang2023creating, ji2023Hierarchical,Ji2023DribbleBot}; (\romannumeral2) manipulation is coarse, lacking precision~\cite{huang2023creating, ji2023Hierarchical}; (\romannumeral3) methods focus on static manipulation, not using the robots' mobile abilities~\cite{cheng2023legmanip, ji2023Hierarchical}. Our approach addresses these issues with a general framework for versatile tasks, precise control, and mobile manipulation.

Utilizing the legs of quadrupedal robots for general manipulation tasks that require a large
workspace and high precision is considerably more complex than merely combining locomotion with manipulation. This complexity introduces several unique challenges.
First, as highly nonlinear systems, loco-manipulators lack the inherent stability of conventional wheeled mobile manipulators. This issue is further exacerbated when legs are used for manipulation and the system is highly underactuated. These characteristics render the problem challenging not only in terms of locomotion but also in manipulation.
Second, the challenge becomes even more complex when incorporating vision for manipulation. Robot vision is crucial for versatile manipulation but is difficult to utilize effectively due to its high dimensionality and the significant gap between simulation and real-world application.

To tackle these challenges, we develop a hierarchical framework that merges Behavior Cloning (BC) with Reinforcement Learning (RL). 
Our framework enables the seamless integration of locomotion and manipulation skills, thereby extending the capabilities of legged robots beyond mere locomotion. The contributions of this work are multifaceted:

\begin{itemize}

\item We carefully design and implement a hierarchical learning framework that harnesses the strengths of both BC and RL, which capitalizes on the efficiency of BC in learning manipulation tasks from demonstrations, as well as the strength of RL in real-time control of high dimensional dynamic systems.

\item Our high-level planner employs a diffusion-based BC policy to efficiently learn a variety of manipulation skills from demonstrations, marking a novel approach in whole-body loco-manipulation.

\item We parameterize the manipulation trajectory of the end-effector for better integration of RL and BC. This method also enables easy data collection through parallel simulations, eliminating the need for teleoperation and the challenges of aligning human actions with legged robots.

\item To evaluate the performance of our algorithm, we design multiple tasks.
These tasks are grounded in practical requirements and are devised to comprehensively evaluate the multifaceted capabilities of loco-manipulators.

\end{itemize}

Collectively, these contributions represent a novel approach to bridging the gap between manipulation and locomotion.

\label{sec:Introduction}

\section{Related Work}
\label{sec:review}


\subsection{Mobile Manipulation}

Traditional mobile manipulator robots typically feature a high-DoF mechanical arm mounted on top of a wheeled chassis~\cite{wu2023tidybot, xiong2024adaptive,fu2024mobile,shafiullah2023bringing,wu2023m,yokoyama2023adaptive,srivastava2022behavior}. 
The low DoF of wheeled platforms lead to robust mobility performance, allowing end-to-end BC to be effective for them in acquiring complex mobile manipulation skills~\cite{shafiullah2023bringing,fu2024mobile}. However, this advantage is counterbalanced by their limitation to flat terrains, which significantly restricts their application scenarios. Legged robots, on the other hand, can readily overcome this limitation.

\subsection{Legged Locomotion}

In recent years, significant advancements have been made in the locomotion capabilities of legged robots. Model-based approaches allow for precise modeling of the robot and environment, enabling robots to achieve robust locomotion skills~\cite{grandia2022perceptive,dtc,ding2021representation}.
Model-free RL has empowered quadrupedal robots to navigate challenging terrains~\cite{cheng2023parkour,zhuang2023robot,hoeller2023anymal,miki2024confined,agarwal2023legged,choi2023deformable}.
Through fine-tuning in the real world, robots can walk in some terrains they have not encountered before~\cite{lei2023uni,smith2022legged}.
By leveraging expert demonstrations to learn motion priors, robots are able to learn various styles of locomotion~\cite{yang2023generalized,wu2023amp,vollenweider2023amp,peng2021amp}.
However, alongside the development of advanced locomotion skills, it is also necessary to cultivate manipulation skills to facilitate their integration into human life.


\subsection{Loco-Manipulation}
Manipulation skills for legged robots have been greatly improved recently. Some researchers choose to augment robots with additional hardware on their backs~\cite{fu2022deep,Versatile2023,gofetch,forrai2023event}, which significantly increases costs and the additional weight compromises the robot's locomotion abilities. Others try to use the robot's legs to perform manipulation tasks. However, these methods have several drawbacks and limitations.
(\romannumeral1) The methods are designed for specific tasks, often restricted to predefined sequences of actions~\cite{cheng2023legmanip, huang2023creating, ji2023Hierarchical,Ji2023DribbleBot}. 
In contrast, we propose a general framework in which a single agent is trained to solve a series of tasks; 
(\romannumeral2) The manipulation is very coarse and does not allow for fine-grained manipulation~\cite{huang2023creating,Ji2023DribbleBot}. 
By contrast, our method can train a low-level controller to achieve precise control for the legs; 
(\romannumeral3) Performing static manipulation, failing to leverage the inherent dynamic capabilities of legged robots~\cite{ji2023Hierarchical}. In contrast, our method enables robots to carry out manipulation tasks while preserving their locomotion abilities; 
(\romannumeral4) It is limited to 3-DoF point tracking, which falls short for more complex manipulation tasks. Additionally, the authors only implemented a low-level controller, necessitating human teleoperation for the execution of all tasks~\cite{arm2024pedipulate}. 
Conversely, our method, utilizing a visual planner, empowers the robot to autonomously execute complex daily manipulation tasks, by tracking both 3-DoF trajectories and rough orientations tracking. 


\section{Hierarchical Learning Framework}
\label{sec:framework}

\begin{figure*}[t]
    \centering
    \includegraphics[width=\linewidth, trim={0 15mm 1cm 0}, clip]{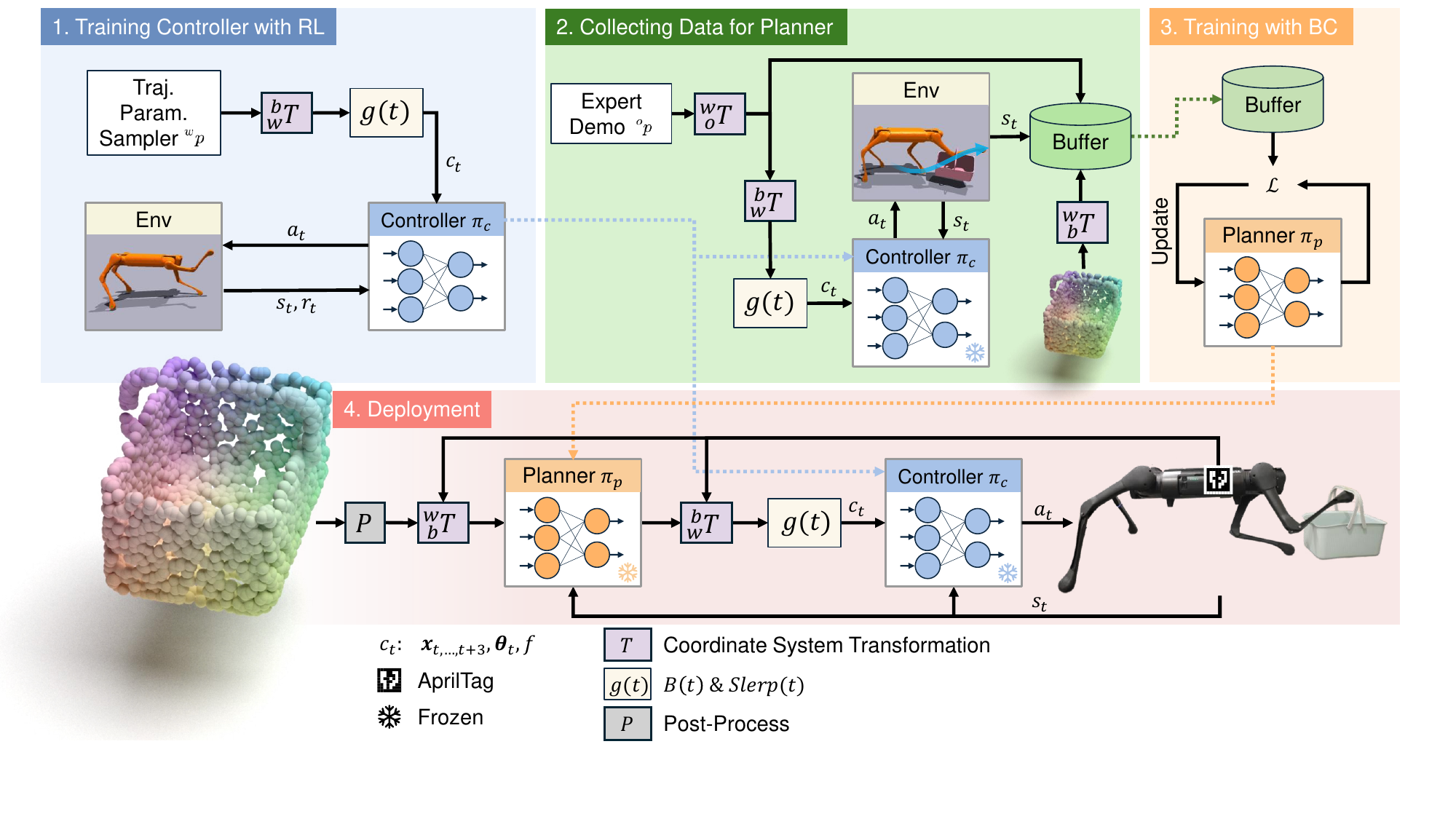}
    \caption{\small (1) We train a control policy $\pi_{c}$ that enables an end-effector to follow curves defined by Bézier control points and weight while maintaining stable locomotion with the other three legs. 
    (2) We use the trained controller to collect expert data. We design manipulation trajectories for different tasks and collect demonstrations through parallel simulation. 
    (3) We use the collected expert data and diffusion-based BC to train the planner.
    (4) In the deployment phase, we use the Realsense D435 to obtain the point cloud, and use an external camera to locate the pose of the robot based on AprilTag~\cite{apriltag2011} for trajectory parameter and point cloud coordinate system transformation. These inputs are sequentially fed into the planner and controller, enabling the robot to perform whole-body loco-manipulation tasks.}
    \label{fig:framework}
    \vspace{-0.5cm}
\end{figure*}

In this section, we introduce our hierarchical learning framework, utilizing BC at the high-level manipulation planning and RL at the low-level joint position control, to enable the robot to achieve versatile loco-manipulation skills.

\subsection{Overview}

As illustrated in Fig.~\ref{fig:framework}, motivated by previous work~\cite{ji2023Hierarchical,huang2023creating}, we decompose the loco-manipulation of the robot into two parts: the high-level planner predicts the desired trajectory parameters for the end-effector, while the low-level controller enables the robot to achieve pose tracking with end-effector. 
The trajectory parameters define the desired pose of the end-effector and identify the legs functioning as manipulators with a manipulator flag $f$.

To develop the high-level planner $\pi_{p}$, we use the trained control policy $\pi_{c}$ to collect expert demonstration data. Specifically, we design expert trajectory parameters for different tasks and make the robot track them in simulation while collecting robot states, point clouds, and trajectory parameters. Through large-scale parallel simulation, we can collect more than 100 expert demonstrations in 3 minutes. Utilizing the collected expert demonstrations, we develop a high-level planner with DP3~\cite{ze2024DP3}, which takes point clouds and robot states as input and outputs trajectory parameters. The frequency of the high-level planner is $10$ Hz.

We develop a low-level controller $\pi_{c}$, which allows the robot to track a 3-DoF target trajectory with any forelimb while maintaining stable walking. This is achieved by training a policy whose inputs include the manipulator flag, the current desired end-effector pose, the current target point of the end-effector calculated by the rational Bézier curve, and the next three target points. The output is desired robot joint positions $q^{d}_{m} \in \mathbb{R}^{12}$. This control policy runs at $50$ Hz, and the desired torque $\mathbf{\tau}$ of the robot is obtained through a PD controller from the desired joint positions.


\subsection{Trajectory Parameterization}
To fully capitalize on the strengths of RL and BC and ensure their effective cooperation, we parameterize the manipulation trajectory of the end-effector. 
These trajectory parameters $p$ serve as both the output of the high-level planner and are utilized to calculate the input for the low-level controller.
Our trajectory parameters include the manipulation flag $f$, 6-order rational Bézier curve parameters, and the target orientation of the start and end points, which we further describe below.

\paragraph{Manipulator Flag} We use any one of the forelimbs as the manipulator, which is specified by a binary manipulation flag $f$, where 0 stands for the Front Left (FL) foot and 1 for the Front Right (FR) foot.

\paragraph{Desired Position Trajectory} In this paper, we use rational Bézier curves to represent the manipulation trajectory. 
Unlike traditional Bézier curves, which are defined solely by control points, rational Bézier curves introduce adjustable weights for these control points, enabling the
 seamless combination of curves and polylines while still maintaining smoothness by weighting the control points. This enables it to parametrically and flexibly represent different trajectories. We consider the foot toe as the end-effector, and the trajectory of the end-effector in 3D space is specified by rational Bézier curves:
\begin{equation}~\label{eq:bezier}
B_{n}(t)=\frac{\sum_{i=0}^n\binom nit^i(1-t)^{n-i}\mathbf{p}_iw_i}{\sum_{i=0}^n\binom nit^i(1-t)^{n-i}w_i}.
\end{equation}
Where $\mathbf{p}_i \in \mathbb{R}^{3\times1}$ represents the control points, $w_i \in \mathbb{R}$ denotes the weights of the control points, introducing an additional degree of control not present in traditional Bézier curves, and $n+1$ is the number of parameters, with $n=6$ chosen for this paper.

\paragraph{Desired Orientation} 
Our manipulation trajectory uniquely specifies the orientation, a capability not realized in previous work~\cite{cheng2023legmanip, ji2023Hierarchical, arm2024pedipulate}. This advancement enables our robot to execute more complex manipulation tasks.

We employ spherical linear interpolation (SLERP) to calculate the desired orientation, which is a method for smooth interpolation between two orientation vectors:
\begin{equation}~\label{eq:Slerp}
Slerp(\mathbf{q}_{0},\mathbf{q}_{1},t)=\frac{\sin[(1-t)\theta]\cdot \mathbf{q}_{0}+\sin(t\theta)\cdot \mathbf{q}_{1}}{\sin\theta}
\end{equation}
Where $\mathbf{q}_{0}$ represents the target orientation of the starting point, and $\mathbf{q}_{1}$ represents the end orientation of the endpoint. $\theta$ is the angle subtended by the arc. 

In Eq.~\eqref{eq:bezier}and\eqref{eq:Slerp}, $t \in [0,1]$ is the phase time that is scaled according to the time span of the trajectory.

\paragraph{The Choice of Reference Frame for Parameters}

Tracking points directly with a mobile robot in RL can face motion asymmetry issues~\cite{2019-MIG-symmetry, Advanced2022Rudin, hoeller2023anymal}, leading us to use the body frame for trajectory parameters ${}^{b}p$ in our low-level control policy. Our high-level planner operates at a lower frequency than the controller, which could cause tracking errors due to time lags between outputs. To prevent this, we use the world frame for both input point clouds ${}^{w}\mathbf{P}$ and output trajectory parameters ${}^{w}p$, avoiding continuous error corrections and swaying motions. Furthermore, when collecting expert data, we randomize robot and object poses, ensuring trajectory parameters ${}^{o}p$ focus on the object by representing them in the object frame and then converting to the world frame based on the pose of the object.

\subsection{Learning Visual Manipulation Planning by BC}
\label{sec:planning}

In this section, we will provide a detailed introduction to the high-level planner, which constitutes the upper layer of the framework. It processes input from point clouds and robot proprioception, and outputs the manipulation trajectory parameters for the end-effector.

\subsubsection{Framework}
Our planner $\pi_{p}$ utilizes DP3~\cite{ze2024DP3} as the backbone, which is a diffusion-based 3D visuomotor policy that can efficiently process 3D data and learn the manipulation trajectory of end-effector from expert demonstrations.

\paragraph{Input}
 We utilize proprioceptive data of the robot state $s_{t}$ and visual point cloud data ${}^{w}\mathbf{P} \in  \mathbb{R}^{n\times 3}$ as inputs. The visual data is captured by a depth camera mounted behind the robot and on its head, which is then transformed into point clouds in the world frame. During the manipulation process, we randomly sample $n=768$ points to form the point clouds.
 
\paragraph{Output}
Our planner generates the parameters ${}^{w}p$ for the manipulation trajectory of the end-effector, represented in the world frame. These parameters can be used to calculate the target point, target orientation, and the manipulator flag $f$ at each moment of the manipulation process. This information is then fed into the subsequent low-level controller.

\subsubsection{Expert Demonstration Collection}

We represent the expert demonstration data as $\begin{aligned}\mathcal{D}=\{\xi_0,\xi_1\ldots\xi_n\}\end{aligned}$, where each trajectory $\xi_i=\{({}^{w}\mathbf{P}_i,s_i,{}^{w}p_i)\}$ is a sequence of point cloud observations ${}^{w}\mathbf{P}$, the robot proprioceptive state $s$, and trajectory parameters ${}^{w}p$.

\begin{figure}[t]
    \centering
    \includegraphics[width=\linewidth, trim={0 92mm 0 0}, clip]{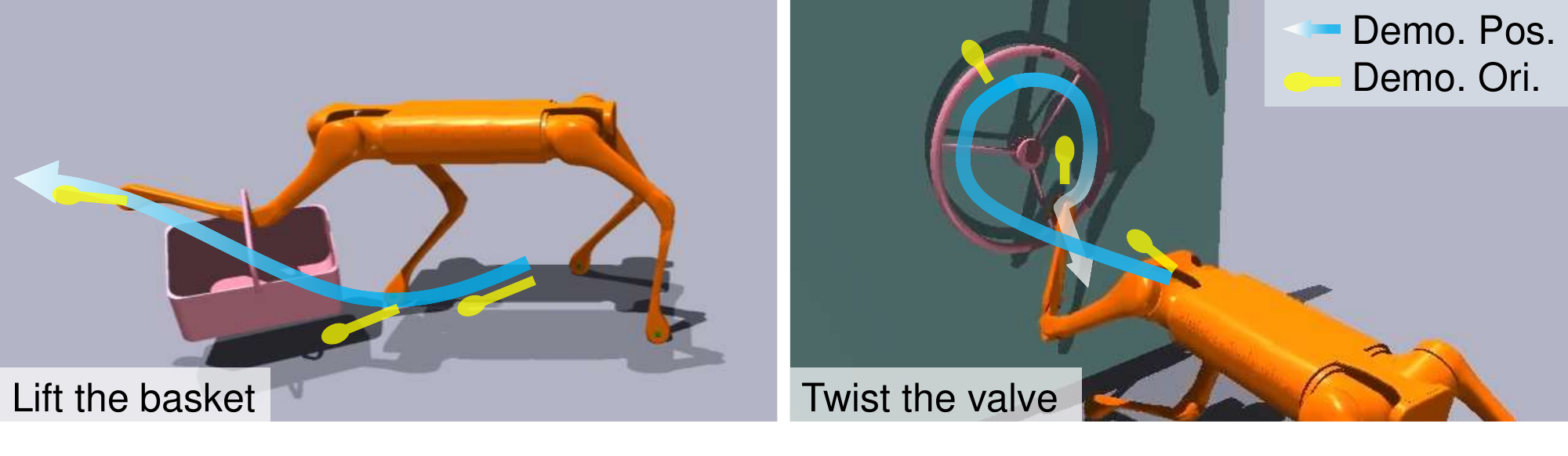}
    \caption{Expert Demonstration. For different objects and tasks, we designed expert trajectory parameters and randomized the poses of the objects. The blue gradient curve represents the demonstration trajectory, and the yellow arrow indicate the demonstration orientation.}
    \label{fig:expert_demo_line}
    \vspace{-.2in}
\end{figure}

As illustrated in Fig.~\ref{fig:expert_demo_line}, we designed expert trajectory parameters ${}^{o}p$ for different objects and tasks. To enable the robot to learn to manipulate objects placed in different poses, we randomized the positions and yaw axis angles of the objects. The expert trajectory parameters are represented in the object frame, allowing them to accurately follow the object. During the collection of expert demonstrations, the object frame is transformed to the world frame based on the object's pose.

We gather expert data by having robots follow expertly designed trajectories, utilizing large-scale parallel simulations in IsaacGym~\cite{makoviychuk2021isaac}. This method enables the rapid collection of a significant volume of data.

The collection time does not increase with the volume of data; it primarily depends on the size of the Video Random Access Memory (VRAM). In this paper, we collect 200 timesteps of expert demonstration data for each task, with 100 trajectories collected per task, taking approximately three minutes.

Despite the collection of expert data in simulations, our method achieves sim-to-real seamless deployment in actual environments, thanks to the point clouds post-process and carefully designed hierarchical structure.

\subsection{Learning Joint Position Control by RL}
\label{sec:control}
In this section, we will introduce the training of the control policy $\pi_{c}$ that enables the robot to track any spatial trajectory with its end-effector while maintaining stable movement with three legs, and the end-effector can handle unknown forces attached to it to ensure the smooth execution of subsequent control tasks. $\pi_{c}$ is trained using the RL algorithm, Proximal Policy Optimization~\cite{schulman2017proximal}.

\subsubsection{Environment}
We train the robot to perform locomotion tasks with three legs, while one leg tracks a given trajectory. The target trajectories are randomly generated within an  $4m\times4m$ square area around the robot, ensuring that the robot learns omnidirectional movement and end-effector tracking tasks. The episode length is 20 seconds, during which the robot is encouraged to track the target trajectories while maintaining stable movement.

\paragraph{Policy Input}
As illustrated in Fig.~\ref{fig:framework}, the input of the control policy is $\mathbf{o}_t$ including past $15$-timesteps history of the proprioception states $\mathbf{s}_t$ and manipulation command $\mathbf{c}_t$. 
The first part is the proprioception states $\mathbf{s}_t$, including the gravity vector in the body frame $\mathbf{g}_{t}$, robot joint position $q_t \in \mathbb{R}^{12}$, robot joint velocity $\dot{q}_t \in \mathbb{R}^{12}$, and action $a_{t-1} \in \mathbb{R}^{12}$.
The second part is the manipulation command $\mathbf{c}_t$ calculated by ~\eqref{eq:bezier} and ~\eqref{eq:Slerp}, including manipulate flag $f$, desired point $\mathbf{x}_t^d \in \mathbb{R}^{3}$, following desired points $\mathbf{x}_{t+1,t+2,t+3}^d$ of the next $3$ timesteps and desired orientation $\theta_t^d$. 
The inputs for both the actor and critic also encompass privileged information, which includes the robot's velocity and the positions of the end-effector as predicted by the state estimator~\cite{estimator}. This estimator takes the same $15$-timesteps history observation as its input.

\paragraph{Action Space}
The action $\mathbf{a}_t$ of the control policy at time step $t$ is the desired joint position $q^d_m \in \mathbb{R}^{12}$. These are passed through a low-pass filter followed by joint-level PD controller to obtain the motor torques $\tau \in \mathbb{R}^{12}$.
\paragraph{Reward}
 We use three types of rewards: a tracking reward for achieving end-effector tracking, a stability term to train the robot's stability, and a smoothness term to ensure smoother movements of the robot.

As shown in Table~\ref{tab:controller-reward}, to precisely follow the target trajectory, our tracking reward includes both position and orientation tracking. Position tracking is determined by comparing the current position of the end-effector with the target position, with the scale in the z-axis direction amplified fivefold to promote leg lifting. Meanwhile, the orientation tracking error is calculated based on the angle difference between the target orientation and the current orientation of the end-effector.
\begin{table}[]
    \scriptsize
    \centering
    \caption{Reward terms for trajectory tracking.}
    \begin{tabular}{ccc}
    \toprule
    \textbf{Term} & \textbf{Expression} & \textbf{Weight} \\
    \midrule
    pos tracking & \( \exp\{{-{|\textbf{x}_{xy}-\textbf{x}^{\text{d}}_{xy}|^2} / {\sigma_{x_{xy}}}}\}\) & \num{0.8} \\
    pos tracking & \( \exp\{{-{|\textbf{x}_{z}-\textbf{x}^{\text{d}}_{z}|^2} / {\sigma_{x_z}}}\}\) & \num{0.8} \\
    ori tracking & \( \exp\{{-(1-(\theta \cdot\theta^{\text{d}})) / {\sigma_{\theta}}}\}\) & \num{0.3} \\
    end-effector accelerations & $\left|\ddot{\textbf{x}}_{ee} \right|^{2}$ & \num{-5} \\
    body accelerations & $\left|\ddot{\textbf{x}}_{base} \right|^{2}$ & \num{-5} \\
    \bottomrule
    \end{tabular}%
    \label{tab:controller-reward}%
    \vspace{-.1in}
\end{table}%

\subsubsection{Domain and Command Randomization}
To address the issue of varying loads on the end-effector during different tasks and the uncertainty of dynamics parameters when deployed in the real world, we randomized the dynamics parameters of the robot and the environment during the training process.

Furthermore, to enable the robot to accurately follow the target trajectory and orientation while moving omnidirectionally and stably, we randomized the manipulation trajectory parameters, as shown in Table~\ref{tab:randomization}.
Note that the Bézier control points are generated within the range of $x,y \in [-2.0, 2.0]$, but this does not limit the robot's range of motion because the Bézier control points will be transformed to be represented in the body frame. This allows the robot to learn to move within an infinite range.

\begin{table}[]
    \centering
    \caption{Randomization range of trajectory parameters.}
    \label{tab:randomization}
    \begin{tabular}{ccc}
    \toprule
    \textbf{Parameter}            & \textbf{Range}                 & \textbf{Unit}    \\ 
    \midrule
    B\'ezier Parameters $\mathbf{p}_{x,y}$   & $[-2.0, 2.0]$ & m       \\
    B\'ezier Parameters $\mathbf{p}_{z}$   & $[0.01, 1.2]$ & m       \\
    B\'ezier Parameters $w$   & $[1, 2000]$ & 1       \\
    Target Orientation $\mathbf{q}_{\phi, \psi}$   & $[0, 2\pi]$ & 1       \\
    Target Orientation $\cos{(\mathbf{q}_{\theta})}$   & $[0.0, 1.0]$ & 1       \\
    \bottomrule
    
    \end{tabular}
    \vspace{-.2in}
\end{table}

\section{Design of tasks for Loco-Manipulation}

\label{sec:tasks}

\begin{figure}[]
    \includegraphics[width=0.99\linewidth, trim={0 17mm 0 0}, clip]{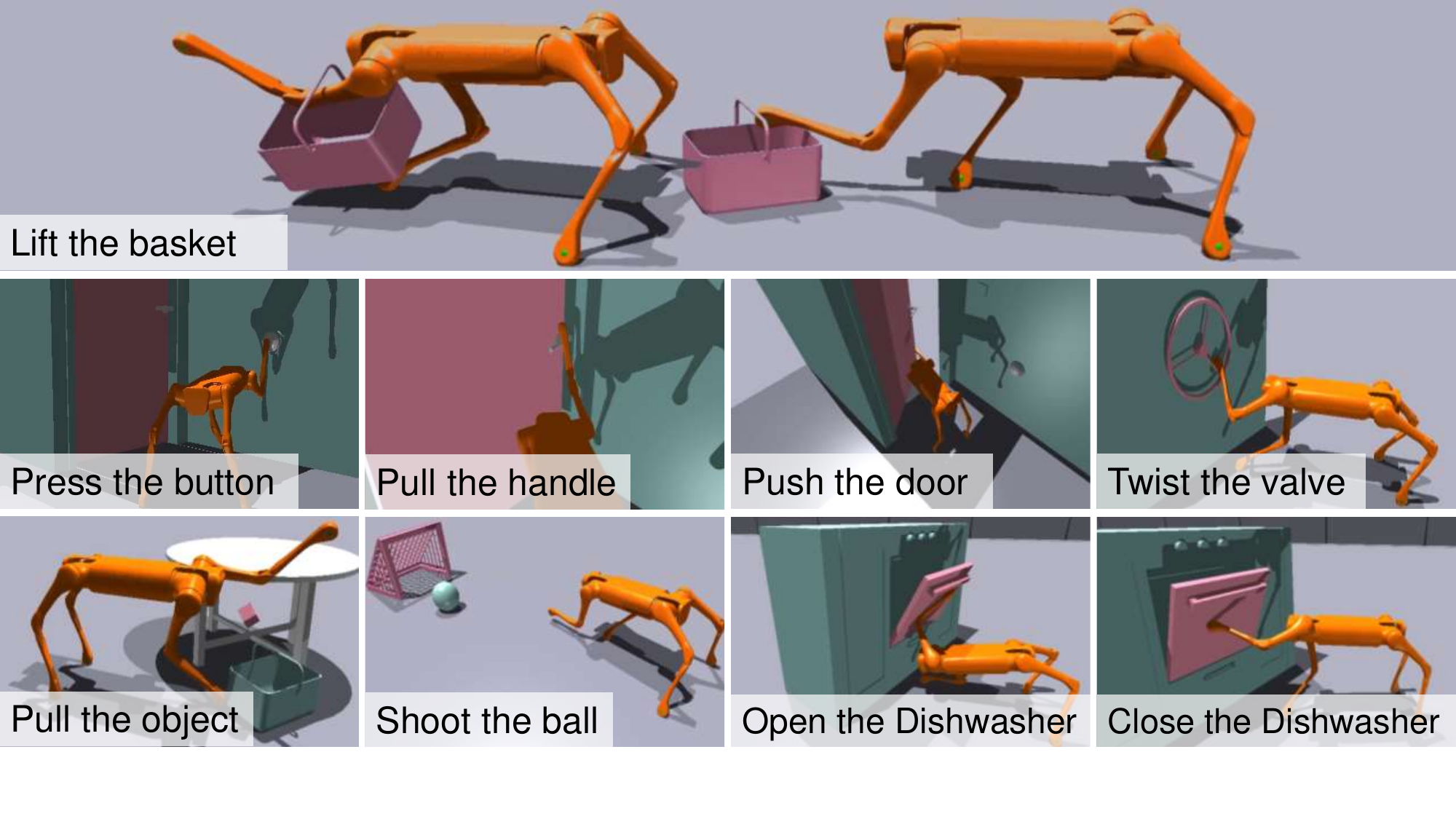}
    \label{fig:basket}
    \caption{\small Overview of the 9 loco-manipulation tasks we train our robot to accomplish. These tasks are designed to cover a large scope of the non-prehensile manipulations tasks that can be realized by the robot's leg.}
    \label{fig:tasks}
    \vspace{-.2in}
\end{figure}
There are numerous robotic manipulation benchmarks available~\cite{yu2019meta,chen2022towards,robosuite2020,RLBench}, but the majority focus on fixed-base robot manipulation. Currently, there is no suitable benchmark for assessing loco-manipulation performance. To evaluate the effectiveness of our algorithm and to provide a valuable reference for the community, we have designed a set of tasks specifically for loco-manipulation tasks, as shown in Fig.~\ref{fig:tasks}.

\subsection{Design Principles}

Based on the application scenarios and working range of legged robots, we designed a set of tasks. Solving these tasks requires the robot to perform various skills, such as pushing, tapping, pulling down, pulling out, kicking, lifting, etc. 
This allows for a comprehensive evaluation of the performance of loco-manipulators across multiple dimensions.
\subsection{Task Description}

\begin{itemize}
    \item \textbf{Press button.} The robot needs to press a hemispherical button with a diameter of 10 cm.
    \item \textbf{Pull handle.} The robot is required to pull down a door handle. Due to the height of the robot's torso and the limitations in joint angles, this task is challenging for loco-manipulation using the feet, requiring the robot to maintain high precision in manipulation at the limits of its operating space.
    \item \textbf{Push door.} The robot needs to push open a door. This assesses the ability of the robot to walk over a large distance on just three legs while manipulating the door with the foot.
    \item \textbf{Lift the Basket.} The robot needs to lift a basket with its foot and walk a distance. This evaluates the robot's precision control over the 6-DoF of its end-effector and its ability to move over a large distance.
    \item \textbf{Open dishwasher.} The robot needs to open the door of a dishwasher with its foot. This task is very easy for a robotic arm with a gripper but very difficult for a foot with point contact. 
    \item \textbf{Close the Dishwasher.} The robot needs to close the door of dishwasher with its foot.
    \item \textbf{Pull objects.} The robot is required to pull objects off a table with its foot.
    \item \textbf{Twist the Valve.} The robot has to twist a valve with a diameter of about 40 cm, where the axis is 60 cm off the ground. This tests the robot's ability to manipulate over a large range at the limits of its workspace.
    \item \textbf{Shoot ball.} The robot needs to run a distance and shoot a soccer ball into a goal with its foot.
\end{itemize}
\section{Experiments}
\label{sec:result}

In this section, we design experiments to test the effectiveness of the proposed method and compare the performance of task completion against different baselines. Subsequently, we validate the learned loco-manipulation skills on an Unitree Aliengo robot and demonstrate sim-to-real transfer capabilities. Finally, we analyze the performance of the proposed method across both the planner and the control policy.

\subsection{Performance Comparison}
\begin{table}[]
\centering
\caption{Performance of our method against baseline Hierarchical Reinforcement Learning (HRL) across 9 tasks. Our method achieves significantly better success rates on all tasks. The success rate for each task is calculated as the average across three seeds. 
}

\begin{tabular}{l|cc}

\toprule
Success rate (\%)    & HRL & \textbf{Ours} \\ 
\midrule
Press Button     & \dd{21.67}{3.51} & \ddbf{92.33}{8.96}  \\
Pull Handle      & \dd{0.00}{0.00} & \ddbf{82.33}{3.21}  \\
Push Door        & \dd{15.67}{3.51} & \ddbf{85.33}{5.03}  \\
Lift Basket      & \dd{11.00}{4.36} & \ddbf{59.33}{12.01} \\
Open Dishwasher  & \dd{1.33}{1.15}  & \ddbf{5.67}{5.51}  \\
Close Dishwasher & \dd{4.67}{1.53} & \ddbf{50.33}{8.14} \\
Pull Objects     & \dd{8.00}{2.65} & \ddbf{12.67}{10.79}\\
Twist Valve      & \dd{10.00}{4.58} & \ddbf{52.33}{10.21}\\
Shoot Football   & \dd{3.67}{2.08}  & \ddbf{26.00}{2.00} \\ \bottomrule
\end{tabular}
\label{tab:success_rate}
\vspace{-.2in}
\end{table}

\begin{table*}[]
\centering
\caption{Comparison of performance on the button pressing and dishwasher closing task. Our method versus end-to-end Behavior Cloning (BC), Hierarchical Reinforcement Learning (HRL) and end-to-end Visual Reinforcement Learning (VRL), each trained with varying amounts of visual data, with the amounts of data labeled after each baseline. M denotes million; K denotes one thousand. Our approach not only achieved a significantly higher success rate but also required considerably less expensive visual data. The success rate is calculated as the average across three seeds. 
}
\begin{tabular}{r|ccccccc}
\toprule
Success rate (\%) & Ours-2K & \textbf{Ours-20K} & BC-20K & BC-500K & HRL-1M & VRL-5M\\ \midrule
Press Button & \dd{42.33}{17.16} & \ddbf{92.33}{8.96} & \dd{0.00}{0.00} & \dd{0.67}{0.58} & \dd{21.67}{3.51} & \dd{0.00}{0.00} \\ 
Close Dishwasher & \dd{33.67}{7.77} & \ddbf{50.33}{8.14} & \dd{0.00}{0.00} & \dd{3.33}{2.08} & \dd{4.67}{1.53} & \dd{0.00}{0.00} \\ 
\bottomrule
\end{tabular}

\label{tab:success_rate_e2ebc}
\vspace{-.2in}
\end{table*}

Our approach utilized 3 billion timesteps of robot state data to train the low-level controller and 20k timesteps of visual data to train the high-level planner. The final success rates for 9 tasks are presented in Table~\ref{tab:success_rate}. 

\begin{itemize}
    \item \textbf{End-to-End BC (BC).} We employ DP3~\cite{ze2024DP3} to train an end-to-end BC policy for the same tasks. During the collection of expert demonstration data, we also gather the outputs of the controller, which means that trajectory $\xi_i=\{({}^{w}P_i,s_i,{}^{w}p_i,a_i)\}$, where $a_i$ is the robot desired joint position.
    \item \textbf{Visual RL as Planner (HRL).} We utilize one of the best visual Reinforcement Learning (RL) algorithms, DrQ-v2~\cite{yarats2021drqv2}, as the planner within our framework. The low-level controller employed is identical to the one integrated within our framework. 
    \item \textbf{End-to-End Visual RL (VRL).} We utilize DrQ-v2~\cite{yarats2021drqv2} to train an end-to-end policy for solving the tasks.
\end{itemize}

Considering the difficulties that most baselines encounter in accomplishing our challenging tasks, we chose to closely examine the button pressing and dishwasher closing tasks. In these tasks, our method achieves the highest success rate and approximately a 50\% success rate, respectively. 
Concurrently, we evaluate the performance of both our method and the HRL baseline across all tasks, which is the best performing baseline on the button pressing and dishwasher closing tasks.

\textbf{Success rate.} 
In Table.~\ref{tab:success_rate}, we compared the performance of our method against HRL across multiple tasks. 
Our method surpasses HRL in all tasks. 
HRL requires meticulous adjustment of rewards for each task; without this, it struggles to learn how to tackle these challenging tasks.

\textbf{Data efficiency.} 
As shown in Table.~\ref{tab:success_rate_e2ebc}, we compare the performance of our method against BC, VRL and HRL for the button pressing and dishwasher closing tasks, testing the performance of each method with an increasing number of visual data. Our method only requires 20k timesteps of visual data to successfully complete the task, while BC and HRL are also difficult to complete the task with a much larger amount of visual data, VRL is completely unable to achieve our task. This data efficiency is achieved by our carefully designed framework, which trains low-level controllers with easily accessible robot state data.

\subsection{Robust Manipulation in Unexpected Situations}
Our expert trajectories are generated in simulation with fixed trajectory parameters, a method that is notably efficient and rapid. However, unlike human video data and teleoperation, this approach does not allow for the collection of data in unexpected situations. Surprisingly, even under these conditions, our method still demonstrated robustness to unforeseen circumstances. 

\begin{figure}[]
    \centering
    \includegraphics[width=0.99\linewidth, trim={0 35mm 16cm 0}, clip]{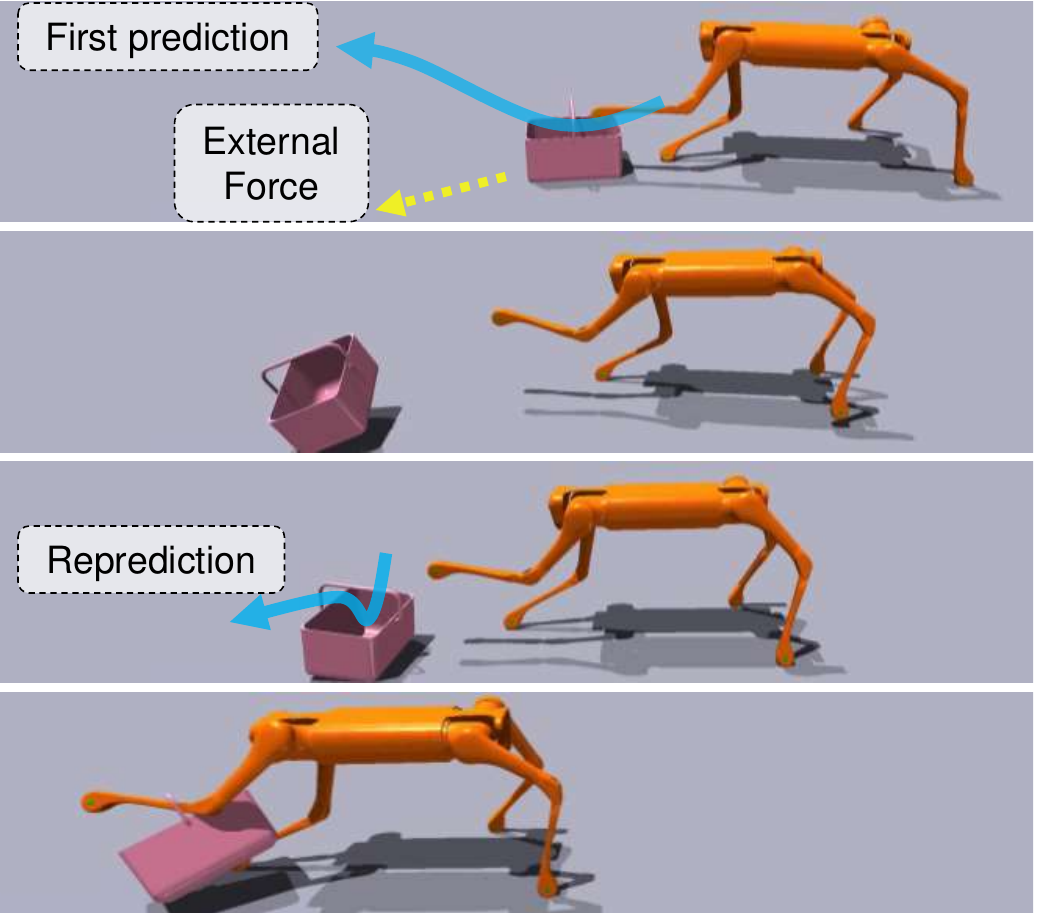}
    \caption{The performance of our proposed method in the task of lifting a basket when encountering unexpected situations. At the start, the robot estimated the trajectory based on initial pose of basket. When the basket was pushed by external force, the robot quickly updated its trajectory prediction to account for the basket's new pose, enabling it to lift the displaced basket successfully.}
    \label{fig:planner_basket}
    \vspace{-.2in}
\end{figure}

Taking the task of lifting a basket as an example, we analyze the robot's manipulation process in the face of unexpected events.
As shown in Fig.~\ref{fig:planner_basket}, at the beginning of the episode, the robot predicted the trajectory parameters according to the basket initial pose. During the approach to the basket, we applied random force to the basket, causing it to roll approximately 1.5 meters away. However, the robot quickly repredicted the trajectory parameters based on the new point clouds of the object, allowing the robot to successfully lift the displaced basket.

\subsection{Sim2real Transfer}

We directly implement the trained controller and planner on a real-world Aliengo quadruped robot without the need for further fine-tuning, demonstrating the robustness of our approach.
As shown in Fig.~\ref{fig:framework}, the pose of the robot is determined using AprilTag for the coordinate system transformation of trajectory parameters and point clouds. Following this, we applied post-process to the point cloud to ensure alignment with the simulation. In Fig.~\ref{fig:teaser}, we illustrate the robot performing tasks like lifting a basket with its forelimbs and pushing the door during locomotion. For demonstrations of all the skills developed through our method, please refer to the website.

\subsection{Planner Performance}

\begin{figure}[t]
    \begin{subfigure}[t]{0.470\linewidth}
    \begin{minipage}[t]{\linewidth}
    \centering
    \includegraphics[width=\linewidth, trim={0 0 8cm 0}, clip]{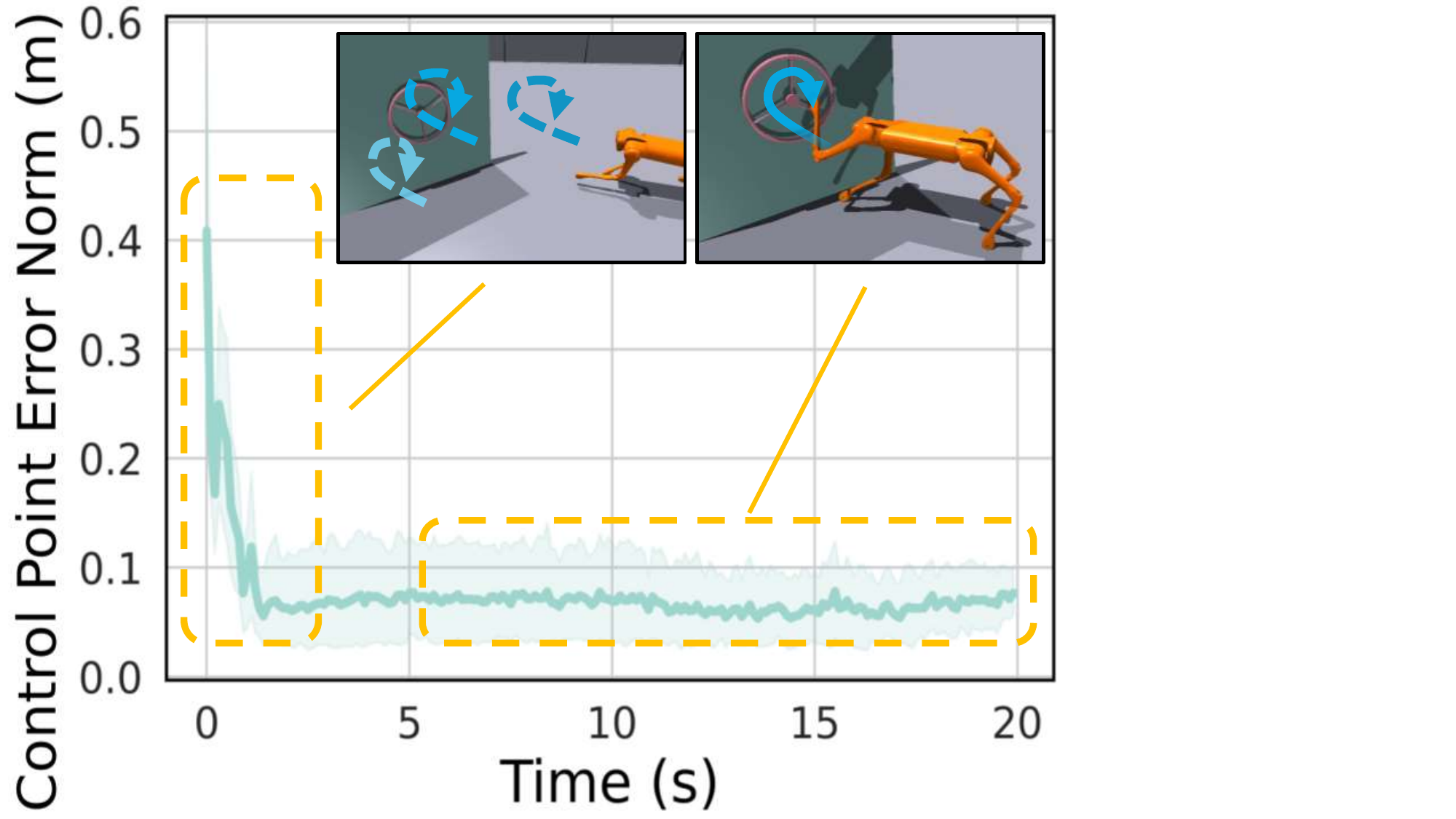}
    \caption{Valve Twisting Task Performance. The green line displays the differences between the predicted parameters and the expert parameters during the manipulation process, as calculated from data collected during 100 tests.}
    \label{fig:planner_valve}
    \end{minipage}
    \end{subfigure}
    \begin{subfigure}[t]{0.510\linewidth}
    \begin{minipage}[t]{\linewidth}
    \centering
    \includegraphics[width=\linewidth]{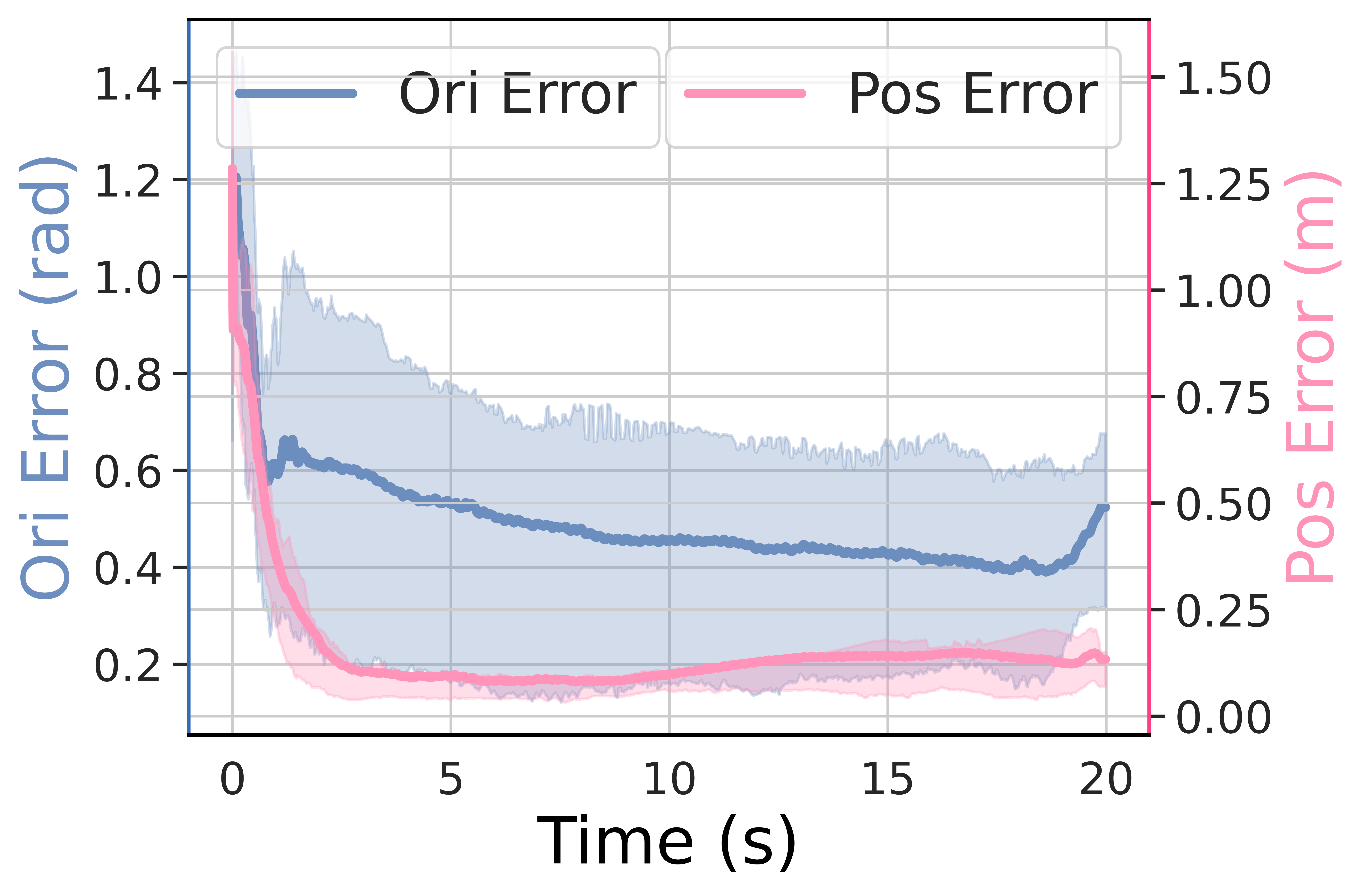}
    \caption{End-effector Trajectory Tracking Performance of the Control Policy. Data on the position and orientation of the end-effector were collected and analyzed during the process of collecting expert demonstrations 10 times for each of the 9 tasks.}
    \label{fig:controller_error}
    \end{minipage}
    \end{subfigure}
    \captionsetup{justification=raggedright,singlelinecheck=false}
    \caption{The predictive and tracking performance of planners and control policy in simulation.}
    \vspace{-.2in}
\end{figure}

The trained planner is capable of predicting the trajectory parameters based on the point clouds and transmitting them to the controller, enabling the robot to complete challenging loco-manipulation tasks, as shown in Fig.~\ref{fig:tasks}.

We conduct an in-depth test of the planner's performance, using the valve twisting task as an example for analysis.
As shown in Fig.~\ref{fig:planner_valve}, at the beginning of the episode, the predicted parameters are constantly changing, leading to a rather chaotic predicted trajectory. As the manipulation progresses, the predicted trajectory tends to converge.

\subsection{Control Policy Performance}

\textbf{Robust tracking against disturbances.}
We test the tracking performance of the control policy. In the tests, the robot is required to follow the expert trajectories for 9 different tasks and record the position and orientation errors of the end-effector in comparison to the desired value. 

As shown in Fig.~\ref{fig:controller_error}, both position and orientation errors peaked when the manipulation began, then position error converged to the lowest value after approximately 3 seconds. In the latter half of the episode, due to the contact with objects, position errors slightly increased but still remained at a low level.
Note that precise tracking of both the position and orientation of the EE is typically unsolvable, given the system is highly underactuated. Our system opts to prioritize position tracking, resulting in the reduction of the minimum orientation error to 0.4 radians. 

\textbf{Teleoperation with control policy.}
\begin{figure}[]
    \centering
    \includegraphics[width=0.495\textwidth, trim={0 5cm 0 0}, clip]{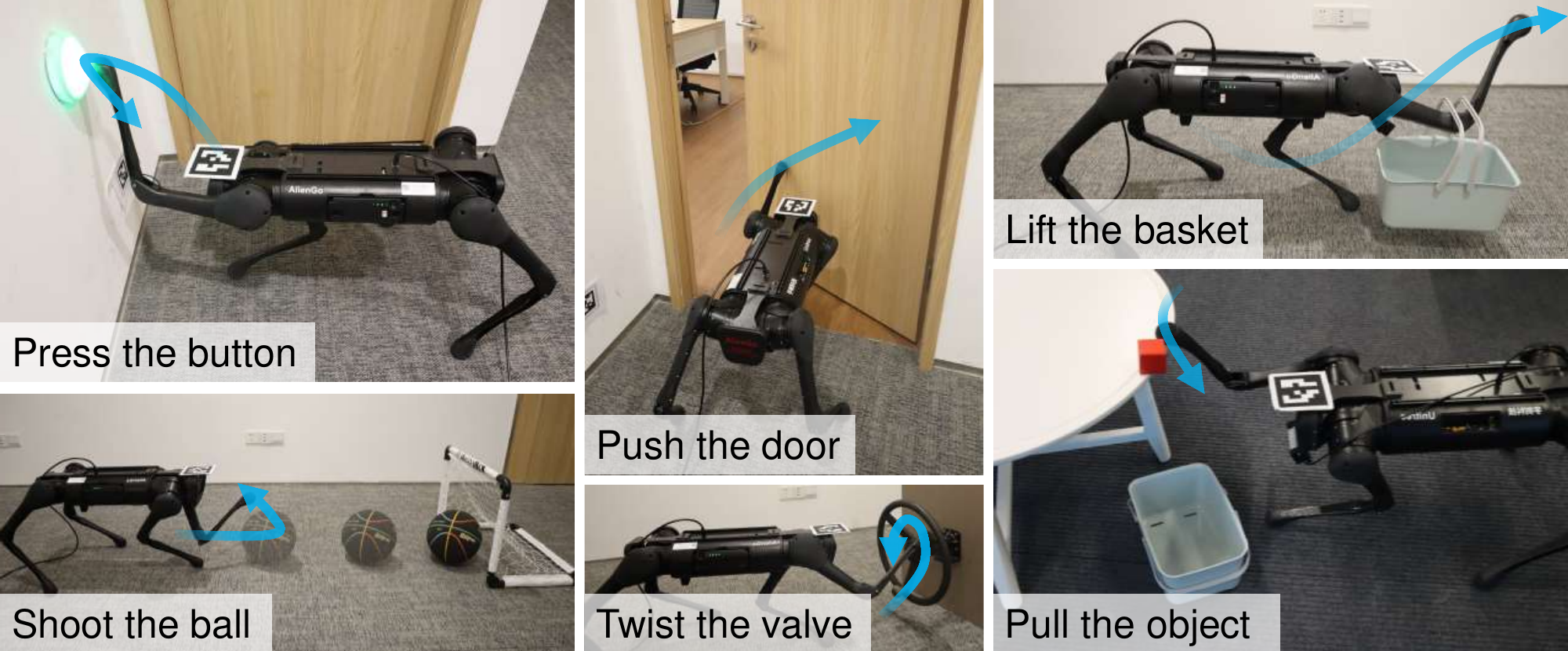}
    \caption{Teleoperation in real world with control policy. We accomplish the above tasks in the real world by the low-level controller with specific trajectory parameters via a joystick.}
    \label{fig:teleop}
    \vspace{-.2in}
\end{figure}
Besides autonomous manipulation, we can \emph{also} collect data via teleopration. As shown in Fig.~\ref{fig:teleop}, using the trained low-level controller, we executed a series of real-world experiments on the robot, guided by specific trajectory parameters. These experiments show that our low-level controller can effectively perform a variety of daily tasks in real-world settings, tasks that previously necessitated a robotic arm.

\section{Conclusion and future works}

In this work, we decompose the loco-manipulation process of legged robots into a low-level controller based on RL and an high-level planner based on BC. 

\paragraph{limitation}
While we have demonstrated the effectiveness of our method in both simulation and real-world scenarios, there are areas for improvement. 
(\romannumeral1) The accumulated gap in both two phases results in relatively poor real-world performance.
(\romannumeral2) While our approach to gathering expert demonstrations is significantly efficient, but we need to post-process the point cloud in deployment to align visual observations.
(\romannumeral3) The inference speed limitations of diffusion-based BC hinder task performance, making it challenging for robots to handle dynamic environments.

\paragraph{Future works}
This study represents a novel effort to master whole-body loco-manipulation, possessing boundless potential. 
(\romannumeral1) It showcases remarkable scalability, enabling rapid collection of expert data for either scale up or data-mixture with real-world data. 
(\romannumeral2) We plan to enhance the planner's inference speed to equip the framework for more dynamic scenarios.




{
\bibliographystyle{IEEEtran}
\bibliography{main}
}

\begin{acronym}
\acro{HP}{high-pass}
\acro{LP}{low-pass}
\end{acronym}

\end{document}